\documentclass[10pt, a4paper]{article}

\usepackage{lrec-coling2024_2} 

\usepackage{tikz}
\usetikzlibrary{trees}

\usepackage{expex}
\usepackage{subcaption}
\usepackage{booktabs}
\usepackage{multirow}
\usepackage{array}
\usepackage{adjustbox}
\usepackage{hhline}
\usepackage{float}
\usepackage{listings}
\usepackage{amsthm}
\usepackage{mdframed}
\usepackage{thmtools}
\theoremstyle{definition}
\declaretheoremstyle[
  headfont=\normalfont\bfseries,
  notefont=\normalfont\bfseries,
  notebraces={}{},
  bodyfont=\footnotesize\itshape,
  postheadspace=1em,
  headpunct={:},
  spaceabove=8pt,
  spacebelow=8pt
]{mystyle}

\declaretheorem[
  style=mystyle,
  name=Example,
]{exmp}

\usepackage{enumitem}
\usepackage{natbib}
\usepackage{multibib}
\makeatletter
\def\@mb@citenamelist{cite,citep,citet,citealp,citealt,citepalias,citetalias}
\makeatother
\newcites{languageresource}{~}

\usepackage{graphicx}
\usepackage{tabularx}
\usepackage{soul}

\usepackage{xcolor}
\usepackage{hyperref}
 \definecolor{darkblue}{rgb}{0, 0, 0.5}
  \hypersetup{colorlinks=true, citecolor=darkblue, linkcolor=darkblue, urlcolor=darkblue}

\usepackage{xstring}

\usepackage{color}

\newcommand{\mattcustomAffiliations}{
\begin{tabular}
{@{}p{\linewidth}@{}}
\\
    \textsuperscript{1}Jožef Stefan Institute $\quad$\textsuperscript{2}Jožef Stefan International Postgraduate School $\quad$ \textsuperscript{4}University of Ljubljana
    \\
\ \ \ \ \ \ \ \ \ \ \ Jamova cesta 39, 1000 Ljubljana, Slovenia  \ \ \ \ \ \ \ \ \ \ \ \ \ \ \ \ \ \ \ \ \ \ \ \ \ \ \ \ \ \ \ \ \ \ \ \ \ \      Ljubljana, Slovenia
    \\
    jaya.caporusso96@gmail.com
    $\quad$
    \{boshko.koloski, senja.pollak\}@ijs.si 
    $\quad$
    mojca.brglez@ff.uni-lj.si
    \\\\
    \end{tabular}\\
    \begin{tabular}
{@{}p{0.5\linewidth}p{0.5\linewidth}@{}}
    $\quad$  $\quad$ $\quad$   $\quad$ \textsuperscript{3}Newcastle University & \textsuperscript{5}Queen Mary University of London \\
   $\quad$  $\quad$ $\quad$ Newcastle upon Tyne, UK & 
   $\quad$  $\quad$  $\quad$ $\quad$  London, UK \\
 $\quad$  $\quad$ d.hoogland2@newcastle.ac.uk 
       & $\quad$  $\quad$ m.purver@qmul.ac.uk \\
  \\     
\end{tabular}
}

\title{A Computational Analysis of the Dehumanisation of Migrants from Syria and Ukraine in Slovene News Media}

\name{
Jaya Caporusso\textsuperscript{1,2*}
$\quad$
Damar Hoogland\textsuperscript{3*}
$\quad$
Mojca Brglez\textsuperscript{4*} 
\\ 
{\bf \large 
Boshko Koloski\textsuperscript{1,2*}
$\quad$
Matthew Purver\textsuperscript{1,5}
$\quad$
Senja Pollak\textsuperscript{1} }
}

\address{\mattcustomAffiliations}

\abstract{
Dehumanisation involves the perception and/or treatment of a social group's members as \textit{less than human}. This phenomenon is rarely addressed with computational linguistic techniques. We adapt a recently proposed approach for English, making it easier to transfer to other languages and to evaluate, introducing a new sentiment resource, the use of zero-shot cross-lingual valence and arousal detection, and a new method for statistical significance testing. We then apply it to study attitudes to migration expressed in Slovene newspapers, to examine changes in the Slovene discourse on migration between the 2015-16 migration crisis following the war in Syria and the 2022-23 period following the war in Ukraine.
We find that while this discourse became more negative and more intense over time, it is less dehumanising when specifically addressing Ukrainian migrants compared to others. 
\\ \\  }

\begin{document}

\maketitleabstract

\section{Introduction}
\label{sec:intro}

Dehumanisation is the perception and/or treatment of a certain social group as if its members were \textit{less than human} \cite{haslam2016recent}. Negating a shared humanity can lead to strong ingroup-outgroup dynamics \cite{arcimaviciene2018migration} and 
discriminatory behaviors \cite{utych2018dehumanization}. 
Dehumanisation and other social biases are reflected and perpetuated through language \cite{hovy2021five} (i.e., \textit{linguistic bias}). While linguistic bias can occur in any modality, this is particularly concerning for text news articles that reach a large audience. 
{\let\thefootnote\relax\footnotetext{$\!\!\!$* These authors contributed equally.}}

During the last decade, migration towards the European Union (EU) has significantly increased on two occasions. In the wake of conflicts in Syria in 2015-16, around 1.3m people entered Europe, and around 7.3m  since the Russian invasion of Ukraine in 2022 
\citep{attitude}\footnote{The people migrating during these two crises were not solely from Syria and Ukraine; we refer to the crises and periods by reference to the locations of their main triggering reasons (the wars in Syria and Ukraine) rather than making any assumptions about the origin or ethnic identity of the people affected.}.
 Although the former crisis 
 was much smaller in scale than the latter, 
it was often represented as a threat to European security  \cite{prideaux2023whitewashing}, while migration from Ukraine was presented in a more welcoming light \citep[e.g.,][]{dravzanova2022attitudes, koppel2023worst, tomczak2023true, zawadzka2023ukrainian}. Such differences in the discourse on migration 
from Syria and Ukraine are often accredited to the higher perceived similarity between ``us'' (typically, EU member state citizens) and the Ukrainian people \cite{bayoumi2022they, pare2022selective}. 

Natural language processing (NLP) methods offer an efficient means to explore relevant phenomena 
including linguistic biases \citep[e.g.][]{NIPS2016_a486cd07}, and dehumanisation \citep[][discussed in detail below]{mendelsohn2020framework}. 
These developments are further driven by pre-trained large language models (LLMs) 
such as BERT \cite{devlin-etal-2019-bert}, GPT-3 \cite{brown2020language}, and XLMR \cite{NEURIPS2019_c04c19c2}. Trained on large datasets, LLMs can successfully capture language structure, and generalise well from few or no examples \cite{brown2020language, wei2021finetuned, kojima2022large}, making them well-suited for knowledge transfer in low-resource settings (in our case, the Slovene language), if task-specific data from high-resourced languages is available. 

Our objective is to characterise changes in Slovenian public attitudes towards migrants, as presented in news, between the Syria and Ukraine migration crisis periods; 
and for the latter period, to describe differences in attitudes 
to Ukrainian and other migrants. To do so, we analyse a corpus of news articles published during these two periods using validated computational methods, here extended and adapted to Slovene. We expect to find the following trends:
\begin{itemize} \item\emph{H1a)} attitudes towards migrants became more positive/intense during the Ukraine period compared to the Syria period; 
\item\emph{H1b)} dehumanising language was more prevalent in the Syria period than 
the Ukraine period; 
\item\emph{H2a)} attitudes towards Ukrainian migrants were more positive/intense than those towards non-Ukrainian migrants; 
and \item\emph{H2b)} dehumanising language was more prevalent in discourse about non-Ukrainian migrants than Ukrainian migrants.
\end{itemize}
Our main contributions are: 
1) adapted computational techniques for analysing dehumanising discourse 
in Slovene, a less-resourced European language;
2) new public resources, including keyword lists for \textit{moral disgust} and \textit{vermin} concepts, and a Valence, Arousal and Dominance (VAD) sentiment lexicon for Slovene;
3) a new method using anchor vectors and the Kolmogorov–Smirnov test to measure
significance of differences in cosine similarities between corpora;
4) an exploration of dehumanisation towards migrants during the Syria and Ukraine migration crises.



This paper is structured as follows. Section~\ref{sec:rel_work} addresses related work. Section~\ref{sec:data} introduces our data and tools. Section~\ref{sec:methods} describes the methods and experimental setup. Results are presented in Sections~\ref{sec:res_periods} and \ref{sec:res_subcorpora}, and discussed, along with conclusion and future work, in Section~\ref{sec:disc}. 

\section{Related work}
\label{sec:rel_work}

While social biases have been widely studied using NLP techniques \cite{liang2021towards}, dehumanisation has only rarely been addressed \cite{wiegand2021implicitly}, perhaps because it is challenging to measure it directly \cite{he2022findings}. However, \citet{mendelsohn2020framework} presented an approach based on five characterizing dehumanisation components \citep{haslam2006dehumanization}: negative evaluation of the target group; denial of agency; moral disgust; likening the target group to something non-human; and psychological distancing and denial of subjectivity. They measured the first two using sentiment detection: employing a lexicon-based approach \cite{mohammad-2018-obtaining_VAD} and connotation frames \cite{rashkin2015connotation}
, they measured valence and dominance, which, together with arousal (i.e., VAD), are considered the three most important dimensions of sentiment \cite{osgood1957measurement}
. Valence represents the continuum between pleasure and displeasure, arousal between engaging and non-engaging, and dominance between control and submission of the experiencer of an affective state \cite{RUSSELL1977273}. To measure the second two elements of dehumanisation, moral disgust and metaphorisation through non-human concepts, \citet{mendelsohn2020framework} employed distributional semantic methods: they measured the cosine similarity of the target group to the concepts of \textit{moral disgust} and \textit{vermin}, showing that this 
can capture interpretable patterns in the discourse on the LGBTQIA+ community in the New York Times; and compared different time periods by constructing and analysing a separate word embedding space for each one.
In this study, we build on their work by modifying and applying it to the discourse on migration in Slovene.

The discourse on migration frequently expresses an ``us'' and ``them'' dichotomy 
\cite{Vezovnik2018SecuritizingMI,vanDijk2018, Chitrakar2020ThreatPI} and portrays migrants as a threat. Migration narratives are also known for their persistent use of mechanistic and animalistic metaphors, equating migrants to water, animals, or commodities \cite{taylor_2021}. 
Recent studies found 
{\it vermin} metaphors to be particularly dominant in anti-immigration online discourse \cite{socsci11080375}. 

\section{Data and resources} 
\label{sec:data}

We base our approach on that of \citet{mendelsohn2020framework}, but address some shortcomings---such as the lack of an arousal analysis and the use of only lexicon-based sentiment analysis---and extend their work in a number of ways. Specifically, we add arousal analysis and neural models for valence and arousal analysis 
(the latter aims to compensate for the possible shortcomings of the lexicon-based approach, as addressed in Section \ref{sec:model})
; introduce a novel inferential anchoring procedure allowing comparison of any two corpora with a shared vocabulary, without the need for vector spaces to be explicitly aligned or share parameters/dimensions; apply the framework to a less-resourced language and a new domain; and, via the novel inferential procedure and cross-lingual valence-arousal model, make it significantly easier to transfer to new datasets.

We first introduce the corpora used 
in Section~\ref{sec:corpora}. In Section~\ref{sec:resources}, we explain the construction of the embeddings model, the construction of the term lists and their concept vectors, and we provide the details of the tools and models used for sentiment analysis.

\subsection{Corpora}
\label{sec:corpora}

We use two corpora of Slovene news, each corresponding to a large-scale migration time period. Both corpora were obtained by one of Slovenia's largest media monitoring companies, and constructed by selecting news articles 
from the online publications of 29 Slovene media outlets. 
The first corpus ($C_{syr}$) contains articles published following the war in Syria and the subsequent migration from August 2015 to April 2016, and the second corpus ($C_{ukr}$) contains articles published during the war in Ukraine from February 2022 to March 2023. 

The corpora were constructed by selecting articles including 
the following migration-related keywords: \textit{begunec*}, \textit{begunc*}, \textit{begunk*}, \textit{beguns*}, \textit{migracij*}, \textit{migrant*}, \textit{imigra*}, \textit{prebežni*}, \textit{pribežni*}, \textit{prebežni*}, and \textit{azil*}. 
These are \textit{unbiased}, almost synonymous terms corresponding to the concepts of \textit{migrant} and \textit{refugee} in English. They were taken from a larger list of migration-related keywords used in previous studies on Slovene \citep[e.g.][]{evkoski_pollak23}, with only the most general terms selected (avoiding, e.g., terms referring to specific nationalities and more loaded terms).




We report descriptive statistics in Table \ref{tab:my_label}.
\begin{table}[H]
    \centering
    \resizebox{0.7\columnwidth}{!}{
    \begin{tabular}{crr}
        \toprule
        Statistic & $C_{ukr}$ & $C_{syr}$~ \\
        \midrule
        documents & 8 470 & 8 556 \\
        sentences & 311 185 & 338 759 \\
        paragraphs & 137 164 & 132 934 \\
        total words & 8 785 219 & 8 282 229 \\
        unique words & 237 622 & 189 512 \\
        total lemmas & 8 785 907 & 8 282 481 \\
        unique lemmas & 100 895 & 77 927 \\
        words per doc. & 1 037.22 & 968.00 \\
        words per sent. & 28.23 & 24.45 \\
        word per par. & 92.19 & 106.95 \\
        \bottomrule
        \end{tabular}
        
        }
    \caption{Dataset statistics.}
    \label{tab:my_label}
\end{table}


Note that the two corpora do not contain news only pertaining to migrants from Syria and Ukraine. In particular, 
in $C_{ukr}$, almost half (49.8\%) of the articles 
do not contain mentions of Ukraine or Ukrainians. For this reason, we further split $C_{ukr}$ into sub-corpora of paragraphs 
(defined as text surrounded by /n/n) mentioning Ukraine ($S_{ukr}$) and paragraphs not mentioning Ukraine ($S_{oth}$), and our analyses include comparisons on the corpus and subcorpus levels. 

\subsection{Analysis Tools}
\label{sec:resources}

 The application of \citet{mendelsohn2020framework}'s method requires the use 
 of a word embedding model, vector representations of selected concepts, and a method to infer sentiment. For the latter, we investigate both lexicon- and classifier-based methods. 

\subsubsection{Static Embeddings}\label{embeddings}
To build a static word embedding model, we first pre-process the corpora to remove titles, and segment text into paragraphs, following \citet{mendelsohn2020framework}. We then apply the CLASSLA tools for South Slavic languages \cite{ljubesic-dobrovoljc-2019-neural} for tokenisation and lemmatisation,
followed by training a Word2Vec model \cite{mikolov2013efficient} for each (sub)corpus. 
To allow for direct comparison of vector spaces, we align the neighbourhoods of the individual models \citep[following][]{kim-etal-2014-temporal} by initialising them from the unlemmatised kontekst.io pre-trained model for Slovene.\footnote{\url{https://kontekst.io/}}
We therefore lemmatise the words in that vocabulary using the word-based LemmaGen3 lemmatiser \cite{jurvsic2010lemmagen} and average the embeddings of any repeated lemmas. 
This reduces our vocabulary by c.58\% 
from \textbf{572,261} word forms to \textbf{242,262} lemmas. 
We train distinct models on the sentences for each corpus 
$C_{syr}$ and $C_{ukr}$. 
Next, we train a model for each subcorpus 
$S_{ukr}$ and $S_{oth}$. Only sentences containing more than two words are considered. We set the min-count to 1 and train the model for 50 epochs.

\subsubsection{Concept vector construction}
\label{sec:conceptvectorconstruction}
We introduce three concept lists that are used to construct the concept embedding vectors, employed in the cosine similarity-based analyses described in Section \ref{sec:methods_vector_similarity}.



\paragraph{Migrant terms}
To select the words forming the \textit{migrant} vector (MV) representing the concept of migrant in the embedding space(s), we start from the list of search terms used to construct the corpora and derive their lemmas (e.g., the search term \textit{migrant*} can capture three lemmas, masculine noun \textit{migrant}, feminine noun \textit{migrantka}, and adjective \textit{migrantski}). For our final concept list, we 
exclude feminine forms as these are rarely present in the corpora and nearly exclusively used in $C_{ukr}$. We also exclude migration-related adjectives and abstract nouns, because these are likely to refer to non-human migration, and can also be used in inherently dehumanising syntagms such as `\textit{migrantski val}' (migrant wave). The final list of the words for the MV is:


\ex<includedVermin>\footnotesize{%
\textit{migrant}, \textit{imigrant}, \textit{begunec}, \textit{azilant}, \textit{prebežnik}, \textit{pribežnik}. (English: \textit{migrant}, \textit{immigrant}, \textit{refugee}, \textit{asylee}, \textit{fugitive},  \textit{escapee}).}
\xe

\paragraph{Moral disgust terms}

 For the \textit{moral disgust} vector (DV), 
 we translate and further select (based on term-frequency analysis) terms identified by \citet{graham2009liberals}.
 Because of the Covid-19 epidemic and the possible effects of its occurrence 
 on the semantics of the models, we exclude disease-related terms. The final list of 76
 terms for the DV includes, among others: 
 
\ex<includedDisgust>\footnotesize{%
\textit{skrunstvo}, \textit{nečist}, \textit{zamazanost}, \textit{prostitut}, \textit{grešnica}, \textit{nezmeren} 
(English: \textit{desecration}, \textit{unclean}, \textit{filthiness}, \textit{prostitute}, \textit{sinner}, \textit{intemperate}
) }
\xe
The complete list of moral disgust terms is available in Appendix \ref{sec:appendixA}.
 
\paragraph{Vermin terms}
Vermin metaphors are a prevalent feature of dehumanising, exclusionary, and racist discourse, and act as the dominant metaphor in offensive anti-immigrant comments \cite{socsci11080375}. We thus also measure dehumanisation through this particular metaphor. To later construct a \textit{vermin} concept vector (VV), we collect vermin-related terms by translating the list of terms used by \citet{mendelsohn2020framework}, based on previous metaphor studies. Our final list of terms is: 
\ex<included>\footnotesize{%
\textit{golazen}, \textit{žužek}, \textit{roj}, \textit{termit}, \textit{parazit}, \textit{zajedavec}, \textit{glodavec}, \textit{miš}, \textit{vampir}, \textit{kobilica}, \textit{ščurek}, \textit{gnida}, \textit{uš}, \textit{pršica}, \textit{bolha}, \textit{pijavka}, \textit{podgana}, \textit{krvoses}, \textit{osa}, \textit{škodljivec}, \textit{mravlja}, \textit{komar}, \textit{žuželka} (English: \textit{vermin}, \textit{bug}, \textit{swarm}, \textit{termite}, 
\textit{parasite}, \textit{rodent}, \textit{mouse}, \textit{bloodsucker}/\textit{vampire}, \textit{locust}, \textit{cockroach}, \textit{louse egg}, \textit{louse}, \textit{mite}, \textit{flea}, \textit{leech}, \textit{rat}, \textit{bloodsucker}, \textit{wasp}, \textit{pest}, \textit{ant}, \textit{mosquito}, \textit{insect})}. 
\xe

\paragraph{Concept vector construction}
Based on these concept lists, we build MV, DV, and VV by taking 
the average of individual word vectors weighted by their frequency in the corpora. 


\subsubsection{Slovene Valence, Arousal and Dominance Lexicon}\label{sec:NRC}

For Slovene, there is no VAD lexicon comparable to the NRC English VAD \citelanguageresource{mohammad-2018-obtaining_VAD} used by \citet{mendelsohn2020framework}. To replicate their valence and dominance analysis, we therefore adapt the English lexicon to Slovene. First, we take the Slovene part of the LiLaH lexicon \citelanguageresource{lilah_daelemans_2020}, a manually validated translation of the NRC Emotion lexicon \citelanguageresource{Mohammad13_emo,mohammad-turney-2010-emotions} containing c.14,000 words with binary values for positive/negative sentiment and 8 basic emotions; for these, we map the English VAD scores directly. 
We then extend this with 5,931 entries not present in LiLaH, translating them using 
sloWNet \citelanguageresource{slownet_fiserv3_1}. 
If no mapping is found, we retain the translation in the machine-translated Slovenian version of the NRC-VAD lexicon. The final resource\footnote{The lexicon is publicly available via CLARIN.SI at 
\url{http://hdl.handle.net/11356/1875}.} contains 19,998 entries with real-valued VAD scores and binary values of sentiment and emotion association. 

\subsubsection{Zero-shot Cross-Lingual VA Detection} 
\label{sec:model}

While the lexicon-based approach above is likely to have high precision, it may have low recall, particularly given the transfer to Slovene. 
Furthermore, it does not capture contextual cues such as word sense, part of speech, and negation \cite{mohammad2020practical}. 
We therefore also use a machine-learning-based approach; given the lack of relevant resources in Slovene, we derive this via cross-lingual transfer of an existing model.
\citet{sentiment_transformer_model} provide 
VAD models 
fine-tuned on 34 datasets from 18 languages (not including Slovene). They investigated three custom losses—Mean Square Error, CCCL, and robust loss—to leverage effective learning through the datasets and RoBERTa family models.  
The best-performing model was \textit{XLM-Roberta-large} \cite{NEURIPS2019_c04c19c2}. 
 The model checkpoint was made available by the authors. We use HuggingFace \cite{wolf-etal-2020-transformers} to infer with the paragraph-level inputs to provide Valence and Arousal (VA). 
We apply this in a zero-shot cross-lingual transfer setting; although the fine-tuning for VA output used no Slovene data, the underlying multilingual language model includes Slovene.

\section{Methods}
\label{sec:methods}
We describe changes in dehumanising attitudes towards migrants expressed in different subcorpora using two methodologies, one based on Word2Vec vector space similarities (Section \ref{sec:methods_vector_similarity}) and the other on sentiment analysis (Section \ref{sec:sent_analysis_meth}).


\subsection{Vector-based similarity analysis} 
\label{sec:methods_vector_similarity}

We use our Word2Vec embeddings (see Section~\ref{embeddings} above) to analyse the differences in the latent representation of the concept of migrant between the corpora. 

\subsubsection {Nearest neighbour analysis}

In each (sub)corpus, 
we first extract the top \textit{k} nearest neighbours (NN) for the MV (see Section \ref{sec:conceptvectorconstruction}), excluding words with the same root as the words used to construct MV. This allows for qualitative inspection of terms (see Section \ref{sec:resNNanalysis}) and functions as input for the sentiment analysis of the corpus-specific NNs lists (see Section \ref{sec:met_hypothesistesting}).

\subsubsection{Similarity of migrants to moral disgust and vermin vectors}\label{sec:methods_distance}

We compare the cosine similarities (CS) of the concept pairs MV-DV and MV-VV in each subcorpus,
and perform a statistical test to assess the significance of the differences in distributions.

\paragraph{Moral disgust} 

Following 
\citet{mendelsohn2020framework}, we assess whether migrants are described with greater or lesser degrees of moral disgust in two corpora by comparing the MV-DV similarity; 
we do this between the two corpora 
%
$C_{ukr}$ and $C_{syr}$, and between the two subcorpora $S_{ukr}$ and $S_{oth}$.

\paragraph{Dehumanising metaphors analysis}  

Similarly, we assess the change in the use of dehumanising metaphorical language by comparing the MV-VV similarity; again we compare both $C_{ukr}$ vs.\ $C_{syr}$ and $S_{ukr}$ vs.\ $S_{oth}$.


\paragraph{Anchoring Kolmogorov-Smirnov test for comparison of neighbourhoods}
To assess whether the difference between corpora in MV-DV or MV-VV similarity is statistically significant, we develop and apply the following novel anchoring procedure, which can be applied without relying on exact alignments between embedding spaces. 


First, we take a selection $ S $ of 1000 random words $ w_i $ from the common vocabulary of the two corpora. 
Next, we use the MV and DV/VV as anchors $v$, denoted by $ v_{mv}$ and $ v_{dv}$, and calculate their distance to each randomly selected word $w_i$ of $S$ as $d(w_i,v)=cos(w_i,v)$, obtaining two vectors that represent each of the two anchors as their distance to each word in $ S $: $a_{mv}=[d(w_1,v_{mv}), d(w_2,v_{mv}),\dots, d(w_N,v_{mv})]$, $a_{dv}=[d(w_1,v_{dv}), d(w_2,v_{dv}), \dots, d(w_N,v_{dv})]$. We then calculate the distance between these two anchor vectors as 
$d=a_{mv}-a_{dv}$. 

We repeat this process for the two corpora to obtain two sets of distances between the anchors, $d_{corpus1}$ and $d_{corpus2}$. 
Finally, we apply the Kolmogorov-Smirnov test 
to assess if $d_{corpus1}$ and $d_{corpus2}$ originated from the same distribution or not, i.e., represent the semantic 
similarities 
between the MV and DV/VV as the same or not. 
We use a conventional $\alpha$=0.05 to draw inferences.

\subsection{Sentiment analysis}
\label{sec:sent_analysis_meth}

We use two approaches to analyse the differences in the sentiment expressed in the corpora.

\subsubsection{Lexicon and transformer approaches}
We apply the lexicon and zero-shot multilingual VA models introduced in Section~\ref{sec:data}. We obtain sentiment scores expressing valence and arousal levels on a scale from 0 to 1 for each paragraph 
in two ways. Following \citet{mendelsohn2020framework}, we use our adapted Slovene VAD lexicon to calculate the score of a paragraph by taking the average 
score over words; here, for each valence and arousal. 
Due to the highly inflectional nature of the Slovene language, 
we use the lemmatised version of the corpus. 
We also employ the ML-based model presented in Section \ref{sec:model}. 
In 
this approach, VA 
scores for each paragraph are predicted from the unlemmatised text. In the initial comparison of the two approaches, we exclude paragraphs with less than 20\% coverage by the NRC lexicon and paragraphs of 15 or fewer words and 500 or more words. We also perform a qualitative analysis of the 20 paragraphs with the highest and lowest scores, to determine which method captures paragraph-level sentiment more successfully. 

\subsubsection{Hypothesis testing}
\label{sec:met_hypothesistesting}

\paragraph{Paragraph-level VA analysis}
To highlight the variations in the attitudes reflected in news reports across the two selected time periods, and between Ukrainian and non-Ukrainian migrants in the second period, we analyse valence and arousal on the paragraph level. We only include paragraphs between 15 and 500 words, with at least five unique words and at least one of the migrant terms (see Section \ref{sec:conceptvectorconstruction}). We compare the VA scores from the method that most accurately describes 
sentiment according to our qualitative analysis. 

\paragraph{Nearest neighbour VAD analysis}
We also look at sentiment on a word level. In each subcorpus, we first extract the top \textit{k} nearest neighbours (NNs) of the MV, disregarding words with the same root as the terms used to construct MV. We compare the 20 NNs of the MV across corpora both qualitatively and quantitatively---the latter, by comparing the sentiment scores (valence, arousal, and dominance) of the 500 NNs using the NRC lexicon.

\paragraph{Bayesian Hypothesis Testing}

We apply Bayesian Hypothesis Testing to assess the difference between distributions of VA 
scores of paragraphs and NN. For each comparison, we adopt normally distributed priors $\mathcal{N}$$(\mu\!=\!0.5,\sigma^{2}\!=\!0.25)$. We assume that the standard deviation of the data is a half-normal distribution with $\sigma^{2}\!\!=\!\!0.25$ and model the data as truncated normal distributions since sentiment scores are defined in $[0,1]$. We use Markov Chain Monte Carlo sampling, drawing 5,000 samples from the posterior distributions after an initial tuning phase of 1,000 samples. We assess modelled probabilities of the difference in means, effect sizes, and credible intervals (CI).


\section{Results: Syria and Ukraine periods }
\label{sec:res_periods}
In this section, we present our comparison of $C_{ukr}$ and $C_{syr}$.

\subsection{Vector-based similarity analysis}\label{sec:similarity_based_results_MEUKR}

\subsubsection{Nearest neighbours analysis} 
\label{sec:resNNanalysis}
We qualitatively analyse the top 20 NNs of each of the $C_{syr}$ and $C_{ukr}$ MVs. The top NNs show a very similar pattern, including \textit{človek}, \textit{tujec}, \textit{oseba}, \textit{prišlek}, \textit{prosilec} (English: \textit{human}, \textit{foreigner}, \textit{person}, \textit{newcomer}, \textit{applicant}). However, while the NNs of MV in $C_{syr}$ contain country names, unique human concepts appear closer to the $C_{ukr}$ vector, such as \textit{sirota}, \textit{pacient}, \textit{ilegalec}, \textit{bolnik}, \textit{družina} (English: \textit{orphan}, \textit{patient}, \textit{illegal}, \textit{patient}, \textit{family}), but also a non-human concept, \textit{žival} (\textit{animal}). 

\subsubsection{Similarity of migrants to moral disgust and vermin vectors}

\paragraph{Moral disgust}
The MV-DV CS 
in $C_{syr}$ (-0.036) is 
lower than in $C_{ukr}$ (0.033). This difference is statistically significant (k=.072, p.011), indicating that \textit{migrant} becomes semantically closer to \textit{moral disgust} in the Ukraine period, which implies a rising trend of migrant dehumanisation. 
This differs from the hypothesised direction (H1b). 


\paragraph{Dehumanising metaphors analysis}
The MV-VV similarity in $C_{ukr}$ (.100) is higher than in $C_{syr}$ (.064). This difference is significant (k=.122, p<.001). Similarly to moral disgust, this indicates that the representation of migrants through vermin-related dehumanising metaphors increases in the Ukraine period, 
again contrary to hypothesis H1b
.


\subsection{Sentiment analysis}
 In Section \ref{VAqualcomparison}, we describe the quantitative and qualitative comparisons of the two sentiment detection approaches. This allows us to select the approach to be used for sentiment analysis and statistical testing of results, as presented in Sections \ref{results_stat_analysis:periods} and \ref{ukr_sub_results_stat_analysis}.

\subsubsection{Lexicon and transformer approaches}\label{sec:sentiment_detection_comparison}
\begin{figure}[h]
    \centering
    \begin{subfigure}{1\linewidth}
        \includegraphics[width=\linewidth]{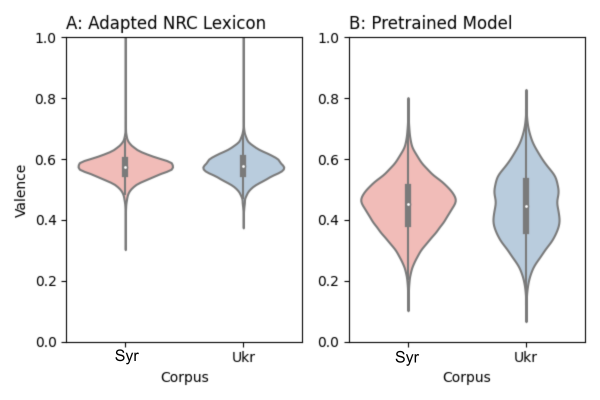}
        \label{fig:violin}
    \end{subfigure}%
    \caption{Distributions of valence scores for $C_{syr}$ and $C_{ukr}$ according to the lexicon approach (A) and the transformer model (B).}
    \label{fig:distribution}
\end{figure}

\paragraph{Quantitative comparison of VA methods}\label{VAquantcomparison}
To compare the two sentiment detection approaches (VAD lexicon and pre-trained cross-lingual model), 
we first compare the overall distributions of VA scores of paragraphs as 
assessed by the two approaches. As illustrated by Figure \ref{fig:distribution}, we note that the valence score distributions obtained using the pre-trained model are wider than the distribution obtained using the lexicon. A similar distribution is observed for arousal scores. The difference in distributions indicates that the scores obtained by the model capture the sentiment expressed in paragraphs in a more fine-grained manner, while the lexicon-based scores all converge around some average score. This global view of VA distributions promotes the use of the pre-trained model over the use of the lexicon.
 
\paragraph{Qualitative analysis of VA methods}\label{VAqualcomparison}
To evaluate the valence scores obtained by the two methods, we also manually evaluate 20 paragraphs per subcorpus, including the top-10 with the highest valence and the top-10 with the lowest valence for each of the two approaches. The evaluation is conducted by a Slovene native speaker. 
They address two aspects of sentiment. In the first step, they assess whether the analysed paragraph presents a positive, negative, or neutral attitude towards migrants (aspect-based sentiment); in the second step, whether the overall sentiment of the paragraph is positive, negative, or neutral (general sentiment). 

Overall, the qualitative evaluation of the highest-valenced (i.e., most positive) paragraphs indicates that both approaches perform well (specifically, paragraphs with a positive valence towards migrants using the lexicon approach: 14/20; paragraphs with a positive valence towards migrants using the crosslingual XLMRoberta approach: 20/20). 
A common theme in these paragraphs is the expression of support, empathy, and solidarity towards migrants, as shown in Example~\ref{ex:example1}.

\begin{exmp}\label{ex:example1}To so torej obrazi ljudi, prostovoljcev, ki nesebično pomagajo beguncem iz dneva v dan in jim s tem vlivajo upanje v nov in boljši jutri. \textcolor{gray}{`So these are the faces of people, volunteers, who selflessly help refugees day by day, giving them hope for a new and better tomorrow.'}
\end{exmp}

In the qualitative analysis of paragraphs with the lowest valence, the picture is a little less clear. Only 9 out of 20 and 7 out of 20 paragraphs for the lexicon-based and model-based methods, respectively, are actually negative towards migrants. In these, a prominent common theme is crimes committed by migrants, including passages that depict migrants in an animalistic manner, arguing for their lack of respect for property, cleanliness, and order.


On the other hand, many of the most negative paragraphs do not necessarily communicate a negative, dehumanising attitude towards migrants. In 3 out of 20 and 7 out of 20 for the lexicon-based and model-based approach, respectively, a common theme is the bad conditions and poor treatment of migrants, and the causes of migration which use negatively-valenced words such as \textit{vojna, slabo, izgubiti, trpljenje} (English: \textit{war, bad, to lose, suffering}). While the topics or events described by the paragraphs are indeed negative, they still communicate a positive attitude towards migrants, accompanied by expressions of support, empathy, and solidarity towards them. 




Although the manual sentiment annotation revealed that migrants are not necessarily negatively evaluated in negatively valenced paragraphs, we find a general trend concerning the dehumanising treatment of migrants. Specifically, while the language is not used to directly dehumanise migrants, it often describes the inhumane and degrading conditions they experience. 
Our qualitative analysis shows that neither resource accurately captures sentiment expressed {\it towards} migrants; however, the neural-model-based approach better captures general sentiment by accounting for the wider context. For these reasons, we use the model predictions in all our subsequent analyses.

\subsubsection{Hypothesis testing}\label{results_stat_analysis:periods}

\paragraph{Paragraph-level VA analysis } 
Hypothesis H1a predicts 
that news about migrants is 
more positive and more intense in the Ukraine period compared to the Syria period, meaning higher valence and arousal scores in $C_{ukr}$ than in $C_{syr}$. This prediction 
is not borne out with regard to valence: by a small margin of .002, valence in $C_{ukr}$ is lower (mean=.446, sd=.108) than in $C_{syr}$ (mean=.449, sd=.089). We find weak evidence that this difference reflects a true population difference, with a probability of .02, a CI close to zero (-.005, -.000), and a posterior Cohen's d of -.024 (CI: -.048, -.001). However, arousal is higher in $C_{ukr}$  (mean=.480, sd=.054) than in $C_{syr}$  (mean=.466, sd=.056) by .013. We find strong evidence that this reflects a robust population difference with a probability of 1.00 (CI: .012, .015) and a posterior Cohen's d of .245 (CI: .223, .267).


\paragraph{Nearest neighbours VAD analysis}
We compare the sentiment scores between the 500 NNs of the MV in $C_{ukr}$  and $C_{syr}$ corpora. The NRC lexicon provides only limited coverage of the NNs lists; for the $C_{ukr}$  corpus, only 22.8\% of the words are in the NRC lexicon, and for $C_{syr}$, only 26.4\%. 
\begin{itemize}
\item {\bf Valence - }The NNs of the $C_{ukr}$ MV are higher in valence (mean=.145, sd=.073) than those of the $C_{syr}$ MV (mean=.134, sd=.065) by .010. We find no evidence that this reflects a true population difference, with a high probability of .73, but the CI for this parameter straddling zero (-.018, .032; Posterior Cohen's d: .097 with CI: -.234, .409)

\item {\bf Arousal - }The NNs of the $C_{ukr}$ MV (mean=.125, sd=.046) are higher in arousal than those of the $C_{syr}$ MV (mean=.107, sd=.046) by .018. We find evidence that this reflects a true population difference, with a high probability 1.00 (CI: .006, .033) and a posterior Cohen's d: .399 (CI: .132, .666)

\item {\bf Dominance - }The NNs of the $C_{ukr}$ MV (mean=.134, sd=.055) are higher in dominance than those of the $C_{syr}$ MV (mean=.121, sd=.054) by .013. We find no evidence that this reflects a true population difference, with a high probability at .95 but a CI that straddles zero (-.002, .030) and a posterior Cohen's d: .239 (CI: -.040, .512).
\end{itemize}


\section{Results: Ukraine sub-corpora}\label{sec:res_subcorpora}
In this section, we present our comparison of the subcorpora of news articles produced during the Ukrainian migration crisis—including, respectively, articles mentioning ($S_{ukr}$) and not mentioning ($S_{oth}$) Ukraine.

\subsection{Vector-based similarity analysis}
\subsubsection{Nearest neighbours analysis}
The top 10 NNs of the two MVs in $S_{ukr}$ and $S_{oth}$ have slightly different orders of similarity. We first ensure that each subcorpus corresponds to a different nationality group by verifying that \textit{Ukrainian} appears only in the $S_{ukr}$  MV neighbourhood and \textit{Kurd} appears only in the  $S_{oth}$ MV neighbourhood. Second, while the first two NNs of MV in $S_{ukr}$  are \textit{človek} (\textit{human}) and \textit{prosilec} (\textit{applicant}), the terms closer to MV in $S_{oth}$ are \textit{priseljenec} (\textit{immigrant}, \textit{settler}) and \textit{tujec} (\textit{foreigner}), implying that media frames Ukrainian migrants as less foreign or alien compared to other nationalities.
Moreover, NNs of MV in $S_{ukr}$ present a higher number of human roles (e.g., \textit{otrok}, \textit{sirota}, \textit{učenec}, \textit{študent}; English: \textit{child}, \textit{orphan}, \textit{pupil}, \textit{citizen}, \textit{student}), while NNs of MV in $S_{oth}$ include more impersonal roles \textit{potnik}, \textit{civilist}, \textit{ilegalec} (\textit{passenger}/\textit{traveler}, \textit{civilian}, \textit{illegal}). However, the top 10 NNs of the  $S_{ukr}$ MV do include a very dehumanising term: \textit{žival} (animal).


\subsubsection{Similarity of migrants to moral disgust
and vermin vectors}
\paragraph{Moral disgust} 
The MV-DV CS in $S_{oth}$ (.068) is larger than in $S_{ukr}$ (.038). This difference is significant (k=.095, p<.001). This result indicates that news articles published during the period of the war in Ukraine communicate less moral disgust 
when they address Ukrainian migrants compared to when they address migrants of other nationalities
---as hypothesised (H2b)
. 
\paragraph{Dehumanising metaphors analysis} 
The MV-VV CS in $S_{oth}$ (.056) is smaller than in $S_{ukr}$ (.162). 
This is a non-significant difference (k=.05, p=.164), 
pointing in the opposite direction from what was hypothesised (H2b). Namely, we found no evidence to support the hypothesis that Ukrainian migrants are less associated to dehumanising metaphors than migrants of other nationalities.


\subsection{Sentiment analysis}
\subsubsection{Lexicon and zero-shot approaches}
 Based on our findings in \ref{sec:sentiment_detection_comparison}, we use the pre-trained transformer model to analyse sentiment in $S_{ukr}$ and $S_{oth}$ and omit the lexicon-based results. 

\subsubsection{Hypothesis testing}\label{ukr_sub_results_stat_analysis}
\paragraph{Paragraph-level VA analysis}
Within $C_{ukr}$, we find that the paragraphs in $S_{ukr}$ have a higher valence (mean=.469, sd=.108) than in $S_{oth}$ (mean=.432, sd=.106) by .036. We also find strong evidence that this reflects a population difference, with the probability of the two means being different at 1.00 (CI: .032, .040) and a posterior Cohen's d of .335 (CI: .295, .373). The results support our hypothesis H2a, i.e., that attitudes towards Ukrainian migrants are more positive and intense than those towards other nationalities.

\paragraph{Nearest neighbours VAD analysis} We compare the valence scores between the 500 NNs of the MV in $S_{ukr}$ and $S_{oth}$. As in section~\ref{results_stat_analysis:periods}, the lexicon provides only limited coverage: only 34.4\% of the words from $S_{ukr}$ and 19.6\% of the words from $S_{oth}$ are present in the NRC lexicon. 
Our analyses show that the NNs of $S_{ukr}$ MV are higher in valence and dominance and lower in arousal than those of $S_{oth}$ MV. However, as in Section \ref{results_stat_analysis:periods}, our statistical tests find no notable difference between the two sub-corpora in any of the three sentiment dimensions. 

\section{Discussion and Conclusion}
\label{sec:disc}

We extend 
the framework of 
\citet{mendelsohn2020framework} and apply it to the investigation of dehumanisation of migrants in Slovene news articles in the periods of the 2015-16 and the 2022-23 migration crises. Specifically, we extend their work in the following innovative ways:
adding arousal analysis to the original valence-only approach; 
using neural models for valence and arousal (instead of solely lexicon-based approaches);
introducing a novel inferential anchoring procedure allowing comparison of any two corpora with shared vocabulary, without the need for vector spaces to be explicitly aligned or share parameters/dimensions;
applying the method to a less-resourced language and a new domain;
and, via the novel inferential procedure and cross-lingual valence-arousal model, making the method significantly easier to transfer to new datasets.

Our analysis of linguistic correlates of dehumanisation in the news articles from Syria ($C_{syr}$) and Ukraine ($C_{ukr}$) periods 
show the following. Concerning hypothesis H1a (\textit{``attitudes towards migrants became more positive/intense during the Ukraine period compared to the Syria period''}), contrary to our expectations, valence appears to be significantly higher in $C_{syr}$; this is supported by the paragraph-level but not by the NNs valence analysis, as the latter did not show any significant differences across corpora. However, the arousal appears to be higher for $C_{ukr}$, 
in line with the  part of H1a concerning intensity; 
this is supported by both the paragraph-level and the NNs arousal analysis.  We interpret this pattern as tentative evidence that attitudes towards migrants expressed in news articles have become less positive and more intense during the Ukrainian period.

When looking closer at $C_{ukr}$, we observe that valence and arousal are both significantly higher in paragraphs that mention Ukraine ($S_{ukr}$) than in paragraphs that do not ($S_{oth}$), as put forward by our hypothesis H2a (\textit{``attitudes towards Ukrainian migrants were more positive/intense than those towards non-Ukrainian migrants''}). 


We also observe that the
 mean valence of $S_{oth}$ is lower than that of $C_{syr}$ (.432 and .449 respectively), which is somewhat contrary to the conclusions of \citet{attitude}: namely, that positive attitudes to migrants from Ukraine ``spill over" to attitudes to migrants from other origins.

The hypotheses H1b) (\textit{``dehumanising language was more prevalent in the Syria period compared to the Ukraine period"}) and H2b) (``\textit{dehumanising language was more prevalent in discourse about non-Ukrainian migrants than Ukrainian migrants"}) are analysed from three perspectives. First, in terms of \textbf{Denial of agency}, the NNs dominance analysis does not find any statistically significant difference in any of our corpora, meaning that we do not detect any difference in the degree to which the agency of migrants was denied. However, as \citet{mendelsohn2020framework} also pointed out, this measure of denial of agency is limited in that it does not capture sentence-level information about whose agency is being denied/affirmed. This method may thus be insufficient to detect this aspect of dehumanising language use.
Next, in terms of \textbf{Moral disgust}, contrary to 
H1b), the concept of migrant appears to be closer to the concept of moral disgust in $C_{ukr}$. However, our analyses on the Ukrainian period sub-corpora confirm H2b), as the concept of migrant is closer to the concept of moral disgust in $S_{oth}$. 
Concerning \textbf{Dehumanising metaphor}, contrarily to our expectations (H1b), the concept of vermin is closer to the concept of migrant in $C_{ukr}$.
The higher level of dehumanisation in the later analysed period might relate to increased migratory movements and a context of a general crisis of the EU \cite{bello2022prejudice}. 
At the same time, \citet{schmidt2023political} argue that higher migratory influxes are not related to a more negative perception of migrants, unlike exclusionary discourses by political
elites that are influencing negative attitudes.
The results concerning $S_{oth}$ and $S_{ukr}$, although not statistically significant, showed a tendency in the direction opposite to what was formulated in H2b, suggesting that the concept of vermin
is closer to migrants from Ukraine.

In conclusion, our results show that while news discourse seems to dehumanise migrants more and more, it does so in a selective way. While the general trend of greater dehumanisation holds for migrants in general, Ukrainian migrants are dehumanised to a lesser extent than other migrants. This confirms previous studies' findings (e.g., \citealt{dravzanova2022attitudes}
), supporting that we perceive and treat Ukrainians differently than other migrants due to their higher perceived similarity to ``us" 
\cite{bayoumi2022they, pare2022selective}.

\par In future work, we plan to investigate target-based sentiment analysis. Additionally, we aim to investigate differences in the discourse around different terms; 
in the context of immigration, for example, \citet{zawadzka2023ukrainian} focused on the terms \textit{refugee} and \textit{(im)migrant}, finding the latter ``less legitimising".
Finally, we will apply the dehumanisation analysis framework to investigate the differences between different newspaper sources.


\section*{Data Availability}
The resources, i.e., the list of terms used to build concept vectors and the Slovenian emotion and VAD lexicon, are made available, respectively, in Appendix A and on CLARIN.SI.  

\section*{Acknowledgements}
We acknowledge the financial support from the Slovenian
Research Agency (ARRS) core research programs
Knowledge Technologies (P2-0103) and 'Slovene Language – Basic, Contrastive, and Applied Studies'  (P6-0215),
 and from the projects CANDAS (Computer-assisted multilingual news discourse
analysis with contextual embeddings, No. J6-2581),
SOVRAG (Hate speech in contemporary conceptualizations of
nationalism, racism, gender and migration, No. J5-3102), and EMMA (Slovenian Research and Innovation Agency research project Embeddings-based techniques for Media Monitoring Applications, No. L2-50070). The work of DH was partially supported by Northern Bridge (UK AHRC); MP by Sodestream (UK EPSRC EP/S033564/1); and ARRS supported the work of MB 
within the national research programme 'Slovene Language – Basic, Contrastive, and Applied Studies'  (P6-0215), 
and BK by the Young Researcher Grant (PR-12394). 



\section*{Limitations}
\label{sec:lim}
This paper addresses the dehumanisation of migrants in Slovenian news media using a computational, quantitative methodology. A limitation of our work lies in our reliance on translated English lexis for the study of dehumanisation, which may not perfectly match the context in Slovene. A more comprehensive evaluation of dehumanising elements in language would necessitate the application of qualitative approaches from critical discourse analysis, examining the wider context in texts and discourses. In turn, this could inform new applications of perhaps more suitable quantitative methods. Furthermore, our sentiment detection approaches did not specifically evaluate sentiment towards the target group. This aspect could in the future be addressed with aspect-/target-based sentiment analysis tools. Finally,  the transformer VAD models did not contain any Slovene training data, and therefore their accuracy could be sub-optimal.  




\section{Bibliographical References}\label{sec:reference}

\bibliographystyle{lrec_natbib}
\bibliography{lrec-coling2024-example}

\begin{thebibliography}{5}
\expandafter\ifx\csname natexlab\endcsname\relax\def\natexlab#1{#1}\fi

\bibitem[{Daelemans et~al.(2020)Daelemans, Fi{\v s}er, Franza, Kranj{\v
  c}i{\'c}, Lemmens, Ljube{\v s}i{\'c}, Markov, and Popi{\v
  c}}]{lilah_daelemans_2020}
Walter Daelemans, Darja Fi{\v s}er, Jasmin Franza, Denis Kranj{\v c}i{\'c},
  Jens Lemmens, Nikola Ljube{\v s}i{\'c}, Ilia Markov, and Damjan Popi{\v c}.
  2020.
\newblock \href {http://hdl.handle.net/11356/1318} {The {LiLaH} emotion lexicon
  of croatian, dutch and slovene}.
\newblock Slovenian language resource repository {CLARIN}.{SI}.

\bibitem[{Fi{\v s}er(2015)}]{slownet_fiserv3_1}
Darja Fi{\v s}er. 2015.
\newblock \href {http://hdl.handle.net/11356/1026} {Semantic lexicon of slovene
  {sloWNet} 3.1}.
\newblock Slovenian language resource repository {CLARIN}.{SI}.

\bibitem[{Mohammad(2018)}]{mohammad-2018-obtaining_VAD}
Saif Mohammad. 2018.
\newblock \href {https://doi.org/10.18653/v1/P18-1017} {Obtaining reliable
  human ratings of valence, arousal, and dominance for 20,000 {E}nglish words}.
\newblock In \emph{Proceedings of the 56th Annual Meeting of the Association
  for Computational Linguistics (Volume 1: Long Papers)}, pages 174--184,
  Melbourne, Australia. Association for Computational Linguistics.

\bibitem[{Mohammad and Turney(2010)}]{mohammad-turney-2010-emotions}
Saif Mohammad and Peter Turney. 2010.
\newblock \href {https://aclanthology.org/W10-0204} {Emotions evoked by common
  words and phrases: Using {M}echanical {T}urk to create an emotion lexicon}.
\newblock In \emph{Proceedings of the {NAACL} {HLT} 2010 Workshop on
  Computational Approaches to Analysis and Generation of Emotion in Text},
  pages 26--34, Los Angeles, CA. Association for Computational Linguistics.

\bibitem[{Mohammad and Turney(2013)}]{Mohammad13_emo}
Saif~M. Mohammad and Peter~D. Turney. 2013.
\newblock Crowdsourcing a word-emotion association lexicon.
\newblock \emph{Computational Intelligence}, 29(3):436--465.

\end{thebibliography}


\begin{thebibliography}{42}
\expandafter\ifx\csname natexlab\endcsname\relax\def\natexlab#1{#1}\fi

\bibitem[{Arcimaviciene and Baglama(2018)}]{arcimaviciene2018migration}
Liudmila Arcimaviciene and Sercan~Hamza Baglama. 2018.
\newblock Migration, metaphor and myth in media representations: The
  ideological dichotomy of “them” and “us”.
\newblock \emph{Sage Open}, 8(2):2158244018768657.

\bibitem[{Bayoumi(2022)}]{bayoumi2022they}
Moustafa Bayoumi. 2022.
\newblock They are ‘civilised’and ‘look like us’: the racist coverage
  of ukraine.
\newblock \emph{The Guardian}, 2.

\bibitem[{Bello(2022)}]{bello2022prejudice}
Valeria Bello. 2022.
\newblock Prejudice and cuts to public health and education: A migration crisis
  or a crisis of the european welfare state and its socio-political values?
\newblock \emph{Societies}, 12(2):51.

\bibitem[{Bolukbasi et~al.(2016)Bolukbasi, Chang, Zou, Saligrama, and
  Kalai}]{NIPS2016_a486cd07}
Tolga Bolukbasi, Kai-Wei Chang, James~Y Zou, Venkatesh Saligrama, and Adam~T
  Kalai. 2016.
\newblock \href
  {https://proceedings.neurips.cc/paper/2016/file/a486cd07e4ac3d270571622f4f316ec5-Paper.pdf}
  {Man is to computer programmer as woman is to homemaker? debiasing word
  embeddings}.
\newblock In \emph{Advances in Neural Information Processing Systems},
  volume~29. Curran Associates, Inc.

\bibitem[{Brown et~al.(2020)Brown, Mann, Ryder, Subbiah, Kaplan, Dhariwal,
  Neelakantan, Shyam, Sastry, Askell, Agarwal, Herbert-Voss, Krueger, Henighan,
  Child, Ramesh, Ziegler, Wu, Winter, Hesse, Chen, Sigler, Litwin, Gray, Chess,
  Clark, Berner, McCandlish, Radford, Sutskever, and
  Amodei}]{brown2020language}
Tom~B. Brown, Benjamin Mann, Nick Ryder, Melanie Subbiah, Jared Kaplan,
  Prafulla Dhariwal, Arvind Neelakantan, Pranav Shyam, Girish Sastry, Amanda
  Askell, Sandhini Agarwal, Ariel Herbert-Voss, Gretchen Krueger, Tom Henighan,
  Rewon Child, Aditya Ramesh, Daniel~M. Ziegler, Jeffrey Wu, Clemens Winter,
  Christopher Hesse, Mark Chen, Eric Sigler, Mateusz Litwin, Scott Gray,
  Benjamin Chess, Jack Clark, Christopher Berner, Sam McCandlish, Alec Radford,
  Ilya Sutskever, and Dario Amodei. 2020.
\newblock Language models are few-shot learners.
\newblock In \emph{Proceedings of the 34th International Conference on Neural
  Information Processing Systems}, NIPS'20, Red Hook, NY, USA. Curran
  Associates Inc.

\bibitem[{Chitrakar(2020)}]{Chitrakar2020ThreatPI}
Rok Chitrakar. 2020.
\newblock \href {https://api.semanticscholar.org/CorpusID:231718116} {Threat
  perception in online anti-migrant speech: a slovene case study}.
\newblock \emph{Sloven{\v{s}}{\v{c}}ina 2.0: empirical, applied and
  interdisciplinary research}.

\bibitem[{Conneau and Lample(2019)}]{NEURIPS2019_c04c19c2}
Alexis Conneau and Guillaume Lample. 2019.
\newblock \href
  {https://proceedings.neurips.cc/paper_files/paper/2019/file/c04c19c2c2474dbf5f7ac4372c5b9af1-Paper.pdf}
  {Cross-lingual language model pretraining}.
\newblock In \emph{Advances in Neural Information Processing Systems},
  volume~32. Curran Associates, Inc.

\bibitem[{Devlin et~al.(2019)Devlin, Chang, Lee, and
  Toutanova}]{devlin-etal-2019-bert}
Jacob Devlin, Ming-Wei Chang, Kenton Lee, and Kristina Toutanova. 2019.
\newblock \href {https://doi.org/10.18653/v1/N19-1423} {{BERT}: Pre-training of
  deep bidirectional transformers for language understanding}.
\newblock In \emph{Proceedings of the 2019 Conference of the North {A}merican
  Chapter of the Association for Computational Linguistics: Human Language
  Technologies, Volume 1 (Long and Short Papers)}, pages 4171--4186,
  Minneapolis, Minnesota. Association for Computational Linguistics.

\bibitem[{Dra{\v{z}}anov{\'a} and Geddes(2022)}]{dravzanova2022attitudes}
Lenka Dra{\v{z}}anov{\'a} and Andrew Geddes. 2022.
\newblock Attitudes towards ukrainian refugees and governmental responses in 8
  european countries.
\newblock In \emph{EU Responses to the Large-Scale Refugee Displacement from
  Ukraine: An Analysis on the Temporary Protection Directive and Its
  Implications for the Future EU Asylum Policy}, pages 135---147. European
  University Institute.

\bibitem[{Evkoski and Pollak(2023)}]{evkoski_pollak23}
Bojan Evkoski and Senja Pollak. 2023.
\newblock \href {https://doi.org/10.14746/amup.9788323241775} {\emph{XAI in
  Computational Linguistics: Understanding Political Leanings in the Slovenian
  Parliament}}, pages 56--61.

\bibitem[{Graham et~al.(2009)Graham, Haidt, and Nosek}]{graham2009liberals}
Jesse Graham, Jonathan Haidt, and Brian~A Nosek. 2009.
\newblock Liberals and conservatives rely on different sets of moral
  foundations.
\newblock \emph{Journal of personality and social psychology}, 96(5):1029.

\bibitem[{Haslam(2006)}]{haslam2006dehumanization}
Nick Haslam. 2006.
\newblock Dehumanization: An integrative review.
\newblock \emph{Personality and social psychology review}, 10(3):252--264.

\bibitem[{Haslam and Stratemeyer(2016)}]{haslam2016recent}
Nick Haslam and Michelle Stratemeyer. 2016.
\newblock Recent research on dehumanization.
\newblock \emph{Current Opinion in Psychology}, 11:25--29.

\bibitem[{He et~al.(2022)He, Ji, Liu, Li, Chang, Poria, Lin, Buntine, Liakata,
  Yan et~al.}]{he2022findings}
Yulan He, Heng Ji, Yang Liu, Sujian Li, Chia-Hui Chang, Soujanya Poria,
  Chenghua Lin, Wray Buntine, Maria Liakata, Hanqi Yan, et~al. 2022.
\newblock Findings of the association for computational linguistics:
  Aacl-ijcnlp 2022.
\newblock In \emph{Findings of the Association for Computational Linguistics:
  AACL-IJCNLP 2022}.

\bibitem[{Hovy and Prabhumoye(2021)}]{hovy2021five}
Dirk Hovy and Shrimai Prabhumoye. 2021.
\newblock Five sources of bias in natural language processing.
\newblock \emph{Language and Linguistics Compass}, 15(8):e12432.

\bibitem[{Jur{\v{s}}ic et~al.(2010)Jur{\v{s}}ic, Mozetic, Erjavec, and
  Lavrac}]{jurvsic2010lemmagen}
Matjaz Jur{\v{s}}ic, Igor Mozetic, Tomaz Erjavec, and Nada Lavrac. 2010.
\newblock Lemmagen: Multilingual lemmatisation with induced ripple-down rules.
\newblock \emph{Journal of Universal Computer Science}, 16(9):1190--1214.

\bibitem[{Kim et~al.(2014)Kim, Chiu, Hanaki, Hegde, and
  Petrov}]{kim-etal-2014-temporal}
Yoon Kim, Yi-I Chiu, Kentaro Hanaki, Darshan Hegde, and Slav Petrov. 2014.
\newblock \href {https://doi.org/10.3115/v1/W14-2517} {Temporal analysis of
  language through neural language models}.
\newblock In \emph{Proceedings of the {ACL} 2014 Workshop on Language
  Technologies and Computational Social Science}, pages 61--65, Baltimore, MD,
  USA. Association for Computational Linguistics.

\bibitem[{Kojima et~al.(2022)Kojima, Gu, Reid, Matsuo, and
  Iwasawa}]{kojima2022large}
Takeshi Kojima, Shixiang~(Shane) Gu, Machel Reid, Yutaka Matsuo, and Yusuke
  Iwasawa. 2022.
\newblock Large language models are zero-shot reasoners.
\newblock In \emph{Advances in Neural Information Processing Systems},
  volume~35, pages 22199--22213.

\bibitem[{Koppel and Jakobson(2023)}]{koppel2023worst}
Katrina Koppel and Mari-Liis Jakobson. 2023.
\newblock Who is the worst migrant? migrant hierarchies in populist
  radical-right rhetoric in estonia.
\newblock In \emph{Anxieties of Migration and Integration in Turbulent Times},
  pages 225--241. Springer International Publishing Cham.

\bibitem[{Liang et~al.(2021)Liang, Wu, Morency, and
  Salakhutdinov}]{liang2021towards}
Paul~Pu Liang, Chiyu Wu, Louis-Philippe Morency, and Ruslan Salakhutdinov.
  2021.
\newblock Towards understanding and mitigating social biases in language
  models.
\newblock In \emph{International Conference on Machine Learning}, pages
  6565--6576. PMLR.

\bibitem[{Ljube{\v{s}}i{\'c} and
  Dobrovoljc(2019)}]{ljubesic-dobrovoljc-2019-neural}
Nikola Ljube{\v{s}}i{\'c} and Kaja Dobrovoljc. 2019.
\newblock \href {https://doi.org/10.18653/v1/W19-3704} {What does neural bring?
  analysing improvements in morphosyntactic annotation and lemmatisation of
  {S}lovenian, {C}roatian and {S}erbian}.
\newblock In \emph{Proceedings of the 7th Workshop on Balto-Slavic Natural
  Language Processing}, pages 29--34, Florence, Italy. Association for
  Computational Linguistics.

\bibitem[{Mendelsohn et~al.(2020)Mendelsohn, Tsvetkov, and
  Jurafsky}]{mendelsohn2020framework}
Julia Mendelsohn, Yulia Tsvetkov, and Dan Jurafsky. 2020.
\newblock A framework for the computational linguistic analysis of
  dehumanization.
\newblock \emph{Frontiers in artificial intelligence}, 3:55.

\bibitem[{Mendes and Martins(2023)}]{sentiment_transformer_model}
Gon{\c{c}}alo~A. Mendes and Bruno Martins. 2023.
\newblock Quantifying valence and arousal in text with multilingual
  pre-trained transformers.
\newblock In \emph{{ECIR 2023}: Advances in Information Retrieval}, pages
  84--100, Cham. Springer Nature Switzerland.

\bibitem[{Mikolov et~al.(2013)Mikolov, Chen, Corrado, and
  Dean}]{mikolov2013efficient}
Tomas Mikolov, Kai Chen, Greg Corrado, and Jeffrey Dean. 2013.
\newblock Efficient estimation of word representations in vector space.
\newblock \emph{arXiv preprint arXiv:1301.3781}.

\bibitem[{Mohammad(2020)}]{mohammad2020practical}
Saif~M. Mohammad. 2020.
\newblock \href {http://arxiv.org/abs/2011.03492} {Practical and ethical
  considerations in the effective use of emotion and sentiment lexicons}.

\bibitem[{Moise et~al.(2023)Moise, Dennison, and Kriesi}]{attitude}
Alexandru~D. Moise, James Dennison, and Hanspeter Kriesi. 2023.
\newblock \href {https://doi.org/10.1080/01402382.2023.2229688} {European
  attitudes to refugees after the russian invasion of ukraine}.
\newblock \emph{West European Politics}, 0(0):1--26.

\bibitem[{Osgood et~al.(1957)Osgood, Suci, and
  Tannenbaum}]{osgood1957measurement}
C.E. Osgood, G.J. Suci, and P.H. Tannenbaum. 1957.
\newblock \href {https://books.google.si/books?id=Qj8GeUrKZdAC} {\emph{The
  Measurement of Meaning}}.
\newblock Illini Books, IB47. University of Illinois Press.

\bibitem[{Par{\'e}(2022)}]{pare2022selective}
C{\'e}line Par{\'e}. 2022.
\newblock Selective solidarity? racialized othering in european migration
  politics.
\newblock \emph{Amsterdam Review of European Affairs}, 1(1):42--61.

\bibitem[{Prideaux~de Lacy(2023)}]{prideaux2023whitewashing}
Jade~Marie Prideaux~de Lacy. 2023.
\newblock " the whitewashing of europe": A comparative analysis of migration
  policy towards the middle east and ukraine, as a reflection of european
  identity politics.

\bibitem[{Rashkin et~al.(2015)Rashkin, Singh, and
  Choi}]{rashkin2015connotation}
Hannah Rashkin, Sameer Singh, and Yejin Choi. 2015.
\newblock Connotation frames: A data-driven investigation.
\newblock \emph{arXiv preprint arXiv:1506.02739}.

\bibitem[{Russell and Mehrabian(1977)}]{RUSSELL1977273}
James~A Russell and Albert Mehrabian. 1977.
\newblock \href {https://doi.org/https://doi.org/10.1016/0092-6566(77)90037-X}
  {Evidence for a three-factor theory of emotions}.
\newblock \emph{Journal of Research in Personality}, 11(3):273--294.

\bibitem[{Schmidt-Catran and Czymara(2023)}]{schmidt2023political}
Alexander~W Schmidt-Catran and Christian~S Czymara. 2023.
\newblock Political elite discourses polarize attitudes toward immigration
  along ideological lines. a comparative longitudinal analysis of europe in the
  twenty-first century.
\newblock \emph{Journal of Ethnic and Migration Studies}, 49(1):85--109.

\bibitem[{Taylor(2021)}]{taylor_2021}
Charlotte Taylor. 2021.
\newblock \href {https://doi.org/10.1177/0957926521992156} {Metaphors of
  migration over time}.
\newblock \emph{Discourse \& Society}, 32(4):463--481.

\bibitem[{Tomczak-Boczko et~al.(2023)Tomczak-Boczko, Go{\l}{\k{e}}biowska, and
  G{\'o}rny}]{tomczak2023true}
Justyna Tomczak-Boczko, Klaudia Go{\l}{\k{e}}biowska, and Maciej G{\'o}rny.
  2023.
\newblock Who is a ‘true refugee’? polish political discourse in
  2021--2022.
\newblock \emph{Discourse Studies}, page 14614456231187488.

\bibitem[{Utych(2018)}]{utych2018dehumanization}
Stephen~M Utych. 2018.
\newblock How dehumanization influences attitudes toward immigrants.
\newblock \emph{Political Research Quarterly}, 71(2):440--452.

\bibitem[{van Dijk(2018)}]{vanDijk2018}
Teun~A. van Dijk. 2018.
\newblock \href {https://doi.org/10.1007/978-3-319-76861-8_13} {\emph{Discourse
  and Migration}}, pages 227--245. Springer International Publishing, Cham.

\bibitem[{Vezovnik(2018)}]{Vezovnik2018SecuritizingMI}
Andreja Vezovnik. 2018.
\newblock \href {https://api.semanticscholar.org/CorpusID:152230991}
  {Securitizing migration in slovenia: A discourse analysis of the slovenian
  refugee situation}.
\newblock \emph{Journal of Immigrant \& Refugee Studies}, 16:39 -- 56.

\bibitem[{Wei et~al.(2022)Wei, Bosma, Zhao, Guu, Yu, Lester, Du, Dai, and
  Le}]{wei2021finetuned}
Jason Wei, Maarten~Paul Bosma, Vincent Zhao, Kelvin Guu, Adams~Wei Yu, Brian
  Lester, Nan Du, Andrew~Mingbo Dai, and Quoc~V. Le. 2022.
\newblock \href {https://openreview.net/forum?id=gEZrGCozdqR} {Finetuned
  language models are zero-shot learners}.

\bibitem[{Wiegand et~al.(2021)Wiegand, Ruppenhofer, and
  Eder}]{wiegand2021implicitly}
Michael Wiegand, Josef Ruppenhofer, and Elisabeth Eder. 2021.
\newblock Implicitly abusive language--what does it actually look like and why
  are we not getting there?
\newblock In \emph{Proceedings of the 2021 Conference of the North American
  Chapter of the Association for Computational Linguistics: Human Language
  Technologies}, pages 576--587.

\bibitem[{Wolf et~al.(2020)Wolf, Debut, Sanh, Chaumond, Delangue, Moi, Cistac,
  Rault, Louf, Funtowicz, Davison, Shleifer, von Platen, Ma, Jernite, Plu, Xu,
  Scao, Gugger, Drame, Lhoest, and Rush}]{wolf-etal-2020-transformers}
Thomas Wolf, Lysandre Debut, Victor Sanh, Julien Chaumond, Clement Delangue,
  Anthony Moi, Pierric Cistac, Tim Rault, Rémi Louf, Morgan Funtowicz, Joe
  Davison, Sam Shleifer, Patrick von Platen, Clara Ma, Yacine Jernite, Julien
  Plu, Canwen Xu, Teven~Le Scao, Sylvain Gugger, Mariama Drame, Quentin Lhoest,
  and Alexander~M. Rush. 2020.
\newblock \href {https://www.aclweb.org/anthology/2020.emnlp-demos.6}
  {Transformers: State-of-the-art natural language processing}.
\newblock In \emph{Proceedings of the 2020 Conference on Empirical Methods in
  Natural Language Processing: System Demonstrations}, pages 38--45, Online.
  Association for Computational Linguistics.

\bibitem[{Zawadzka-Paluektau(2023)}]{zawadzka2023ukrainian}
Natalia Zawadzka-Paluektau. 2023.
\newblock Ukrainian refugees in polish press.
\newblock \emph{Discourse \& Communication}, 17(1):96--111.

\bibitem[{Šori and Vehovar(2022)}]{socsci11080375}
Iztok Šori and Vasja Vehovar. 2022.
\newblock \href {https://doi.org/10.3390/socsci11080375} {Reported
  user-generated online hate speech: The `ecosystem', frames, and ideologies}.
\newblock \emph{Social Sciences}, 11(8).

\end{thebibliography}

\section{Language Resource References}
\label{lr:ref}

\bibliographystylelanguageresource{lrec_natbib}
\bibliographylanguageresource{languageresource}

\appendix
\section{List of Moral Disgust Terms}
\label{sec:appendixA}
\textit{skrunstvo, nečist, zamazanost, prostitut, grešnica, nezmeren, zloba, klatež, svetoskrunski, izkoriščevalski, razuzdanost, opolzek, beda, izprijen, perverznež,opolzkost, zamazan, nespodoben, odbijajoč, gnus, brezbožen, vlačuga, razvraten, kužen, profana, prostaški, grešnik, cenenost, svetoskrunskost, izprijenost, kurba, umazanija, grešiti, profan, brezobziren, oskruniti, gnusen, ogaben, prešuštnik, grešen, nečistost, nalezljiv, perverzen, grešenje, skrunjen, ogabnost, oskrunjen, beden, razsipen, nalezljivost, razvrat, umazan, prostaškost, bogokleten, razuzdan, greh, odvraten, okuženost, omadeževanost, kužnost, odvratnost, omadeževan, nespodobnost, profanost, bogokletnost, brezbožnost, prostitucija, cenen, prešuštvo, prešuštnica}

\end{document}